# A Dual-Memory Architecture for Reinforcement Learning on Neuromorphic Platforms

Wilkie Olin-Ammentorp[1,2], Yury Sokolov[1], Maxim Bazhenov[1,2*]

[1] Department of Medicine, University of California, San Diego

[2] Institute for Neural Computation, University of California, San Diego

**\* Correspondence:**

Maxim Bazhenov

mbazhenov@health.ucsd.edu

## Abstract:

Reinforcement learning (RL) is a foundation of learning in biological systems and provides a framework to address numerous challenges with real-world artificial intelligence applications. Efficient implementations of RL techniques could allow for agents deployed in edge-use cases to gain novel abilities, such as improved navigation, understanding complex situations and critical decision making. Towards this goal, we describe a flexible architecture to carry out reinforcement learning on neuromorphic platforms. This architecture was implemented using an Intel neuromorphic processor and demonstrated solving a variety of tasks using spiking dynamics. Our study proposes a usable energy efficient solution for real-world RL applications and demonstrates applicability of the neuromorphic platforms for RL problems.

## Introduction:

As the number of data-collecting devices increases, so too does the need for efficient data processing. Rather than require all data collected from remote devices be processed at a central location, the need for data processing to be performed in-situ is becoming a priority; this is especially true in situations where 'agents' collecting data may need to make critical decisions based on these inputs with low latency (such as in self-driving cars or aerial drones). For such use cases, efficiency of data processing becomes paramount, as energy sources and physical space ('size, weight, and power') come at a premium[1].

Neuromorphic architectures provide one path towards meeting this need. Although there is no universal definition on what constitutes a neuromorphic architecture, these systems generally aim to provide efficient, massively-parallel processing schemes which often use binary 'spikes' to transmit information[2]. Given that a clear definition of a neuromorphic architecture is not yet universally agreed upon, it is difficult to design a single program which can be compiled to any neuromorphic system (as is the case with standard computer architectures). But, by restricting ourselves to the massively-parallel operations which are a common feature of almost all neuromorphic systems, we can create a program which likely be adapted to any platform which meets the emerging definition of what constitutes a neuromorphic system[3]. In this work, we utilize Intel's neuromorphic processor codenamed 'Loihi.'[4]

Reinforcement learning (RL) represents a native way how biological systems learn. Instead of being trained before deployment by massive amounts of labeled data, humans and animals learn continuously from experience by updating a policy based on continuously collected data. This requires learning to occur in-situ rather than depending on slow and costly uploading of new data to a central location where new information would be embedded to the previously trained model, followed by downloading the new model to the agent.

Towards these objectives, we describe a high-level system for carrying out RL tasks which is inspired by several principles of biological computation, particularly by complementary learning systems theory[5], postulating that learning new memories in the brain depends on mutual interaction between cortical and hippocampal networks. We show that such 'dual-memory learner' (DML) can implement methods which can approach an optimal solution to an RL problem. The DML architecture is then implemented in a spiking manner and executed on Intel's Loihi processor. We demonstrate it solving the classic multi-arm bandit problem, as well as more advanced tasks such as navigation through a maze, and the card game Blackjack. To our knowledge, these advanced multi-step problems have not previously been demonstrated being solved solely by a neuromorphic system. We characterize the performance of its current implementation, comment on its characteristics and limitations, and describe improvements which it can undergo in future work.

## Results:

### Dual-memory learner (DML) framework

Monte Carlo (MC) methods provide well-characterized RL techniques for learning optimal policies via episodic experiences; the agent does not need to be equipped with a full model of how the environment will react to its actions in order to learn. Instead, the agent tracks which states it has entered, the actions it has taken, and once an episode concludes, updates its value estimates given its trajectory through the state space. This provides a simple but effective basis for reinforcement learning, and we focus on implementing this method in our architecture (though it can also be extended to more modern temporal difference (TD) and n-step algorithms).[6]

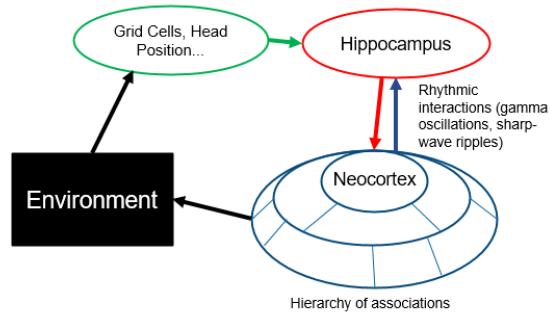

*Figure 1: Biological inspiration for the dual-memory learner.*

We propose a dual-memory learner (DML) framework, which mimics the high-level organization of learning function in the biological brain, so called complimentary learning systems theory (Figure 1)[5] to implement an MC learning technique using spiking networks. The proposed DML architecture contains four major sections which process and store the information required to carry out reinforcement learning on a neuromorphic platform (Figure 2).

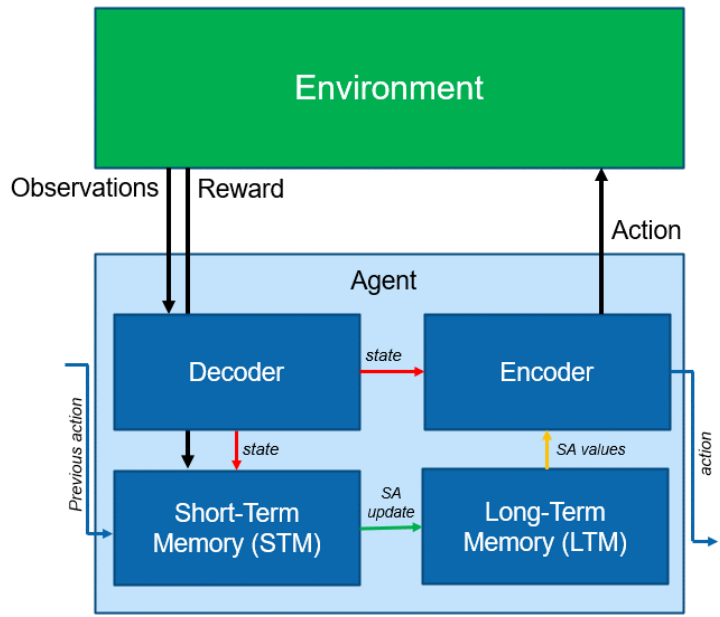

*Figure 2: A high-level overview of the dual-memory learner. Each module is responsible for a sub-step of the evaluation required to choose an action which leads to the next step in a Markov chain process.*

The fundamental epoch or 'step' in an RL agent modeling a Markov decision process requires several sub-steps: recognizing the agent's current state, using this information to decide on an appropriate action given the current policy, returning this action to the environment in a meaningful manner, and possibly applying reward signals to update internal value estimates and policy. We define specific modules and/or interactions to address each of these requirements, forming the core structure of the DML and allowing it to be implemented via parallelized and local operations. The four modules we define are the decoder, short-term memory (STM), long-term memory (LTM), and encoder (Figure 2).

### Architectural Implementation

One of the crucial aspects of a neuromorphic system is the question of how information is to be represented, particularly when all information must be encoded in 'spikes,' the binary all-or-nothing signal which is responsible for almost all information transfer between neurons in the human brain[7]. In this initial implementation, the convention that all information is rate-coded is used. While rate-coding can be costly compared to other encoding strategies and is likely not employed in many regions of the brain[8,9], we use it here due to its easy interpretability and functionality. Using this assumption and following the previously laid-out requirements, we independently demonstrate the operation of each module constituting the DML. With the sole exception of the final encoder module, all modules are implemented entirely using spiking logic which run in the massively parallel 'Neurocores' of the Loihi architecture[4].

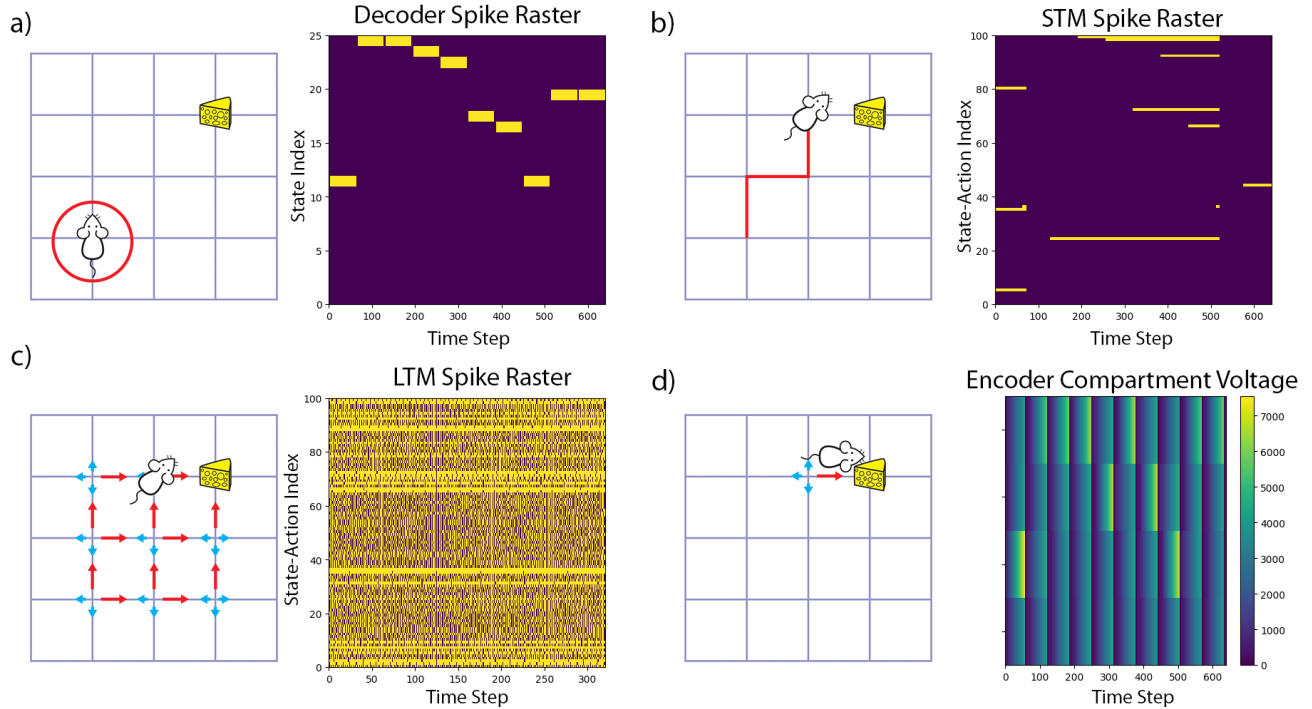

*Figure 3: Overview of each module's purpose and implementation. A) The encoder provides an internal representation of the agent's current state in the task (e.g. the current location of an agent in a maze). B) The short-term memory (STM) stores the current episode's trajectory through state space by storing successive state-action pairs (e.g. previous locations and moves in a maze). C) The long-term memory (LTM) provides estimates of values for each state-action pair in the problem (e.g. the reward expected after moving a certain direction in the maze). D) The encoder chooses an action given the current state and returns it to the environment by reading the appropriate value estimates from the LTM (e.g. finding and taking the cardinal direction in the maze with the highest value).*

### Decoder

In the simplest case, the decoder provides a simple integer value which corresponds to a unique state in the problem. Signals from the environment indicating a change in state are assumed to be sparse in time, and the decoder must take these sparse signals and expand them into a constant internal representation (Figure 3a). This is done by creating a series of bistable neurons which provide a one-hot representation of the current state (e.g. if an agent in a maze is in the fourth possible position, its fourth neuron in the encoder will be active).

### Short-Term Memory (STM)

When combined with a similar one-hot representation of the action taken following the current state, the trajectory of the agent through each episode can be built by taking the outer-product of the decoder's state representation and the consequent action from the encoder (Figure 3b). Again, bistable neurons are used which are activated once a state-action pair has been traversed, and are reset once an episode is complete (e.g. in a maze, an STM neuron becomes active if the agent entered a location and made a specific movement). In all examples presented here, the end of an episode is indicated by a signal from the environment, optionally accompanied by a positive or negative reward.

### Long-Term Memory (LTM)

Within the LTM, the binary trajectory and reward signals must be converted into graded value representations. This is currently done by maintaining a tabular array of value estimates for each state-action pair. This avoids requiring a function approximator, the training and implementation of which is in itself currently an area of intense study within neuromorphic computing[10,11].

To represent a single value estimate, a circuit of several neurons is used. The dynamics in this 'value circuit' (VC) are configured in a manner which allows the output compartment to converge to a firing rate which corresponds to the proportion of reward signals the circuit receives out of all reinforcement signals. Without new reinforcement signals, the VC maintains its current firing rate. The exact details of VC and a proof of its convergence are provided in the materials.

An array of VCs represents the expected returns for all state-action pairs following the current policy (Figure 3c). Agent policy is formed from a simple greedy or ε-greedy choice on action value estimates given the current state. At the end of an episode, these value expectations are updated by using the trajectory stored in the STM to route reward signals to their corresponding VCs. These signals then can incrementally adjust each VC as necessary, creating new value estimates and allowing new policies to be derived.

### Encoder

In order to select an action appropriate to the current state, the encoder uses the information from the decoder to filter the output of the LTM (Figure 3d). Reading only the outputs of the LTM which correspond to actions possible at the current state, the decoder chooses the action with the highest value (greedy), or optionally, may instead choose a random action with a set probability (ε-greedy). Currently, the required argmax and random selection operations are done via the x86 processor co-integrated on the Loihi chip, though they may be replaced via a winner-take-all (WTA) or noisy WTA circuit for a purely spiking implementation of the DML[12].

### Modular Integration

Having demonstrated the individual operation of each DML module, the remaining challenge is to integrate these modules into a system which works in concert to implement the full DML. Additionally, this should be achieved by describing modules at a high level of abstraction, allowing the solution to automatically scale with the problem at hand and preventing it from becoming burdensome to end users who may wish to deploy the program into new scenarios. This remains a challenge in neuromorphic systems where 'completeness' is debatable and compilation of arbitrary programs to end platforms may not always be feasible[3].

We maintain high levels of abstraction in our program integration by describing neuronal circuits and hierarchies of circuits in terms of computational graphs. The elements of these graphs are nodes with arbitrary dimensions, linked with stereotyped connectivity patterns and predefined excitability characteristics. These circuits can then be easily scaled to the given problem at hand and compiled down into the individual compartments and synaptic connections required to define an executable program on a platform such as Loihi. The block diagrams representing these graphs are provided in the methods.

This high-level organization allows the RC-DML to address a variety of different problems while requiring minimal amounts of code updates and maintaining executability on neuromorphic hardware. The only code changes required to allow the agent to address different problems are routines which update the simulation of the environment and control communication between the agent, environment, and host computer. The full source code for the RC-DML on Loihi is openly available (see methods).

### Problem Solving

#### Multi-Arm Bandit

As a basic demonstration of its capabilities, we first apply the rate-coded DML (RC-DML) to the multi-arm bandit (MAB) problem. While it does not incorporate the concept of 'state' into a problem, the MAB is itself a complex problem which addresses many fundamental aspects of RL[6]. In the MAB, a series of 'arms' is presented to the agent, each with a hidden true parameter which controls the probability a reward is obtained when that arm is 'pulled[13].' Here, the fundamental learning problem is to find which arm gives the highest reward using the smallest number of interactions with the bandit in order to maximize cumulative reward.

We demonstrate RC-DML using an ε-greedy algorithm to solve the MAB. As the problem is stateless, it is simply indicated to the system that it remains in the same (singular) state after every action. The value estimates learned for each action in this state are then used to estimate the reward for each arm (Figure 4a). Using an ε-greedy policy, the RC-DML is forced to explore each arm and eventually converges on selecting the correct arm (Figure 4b, c).

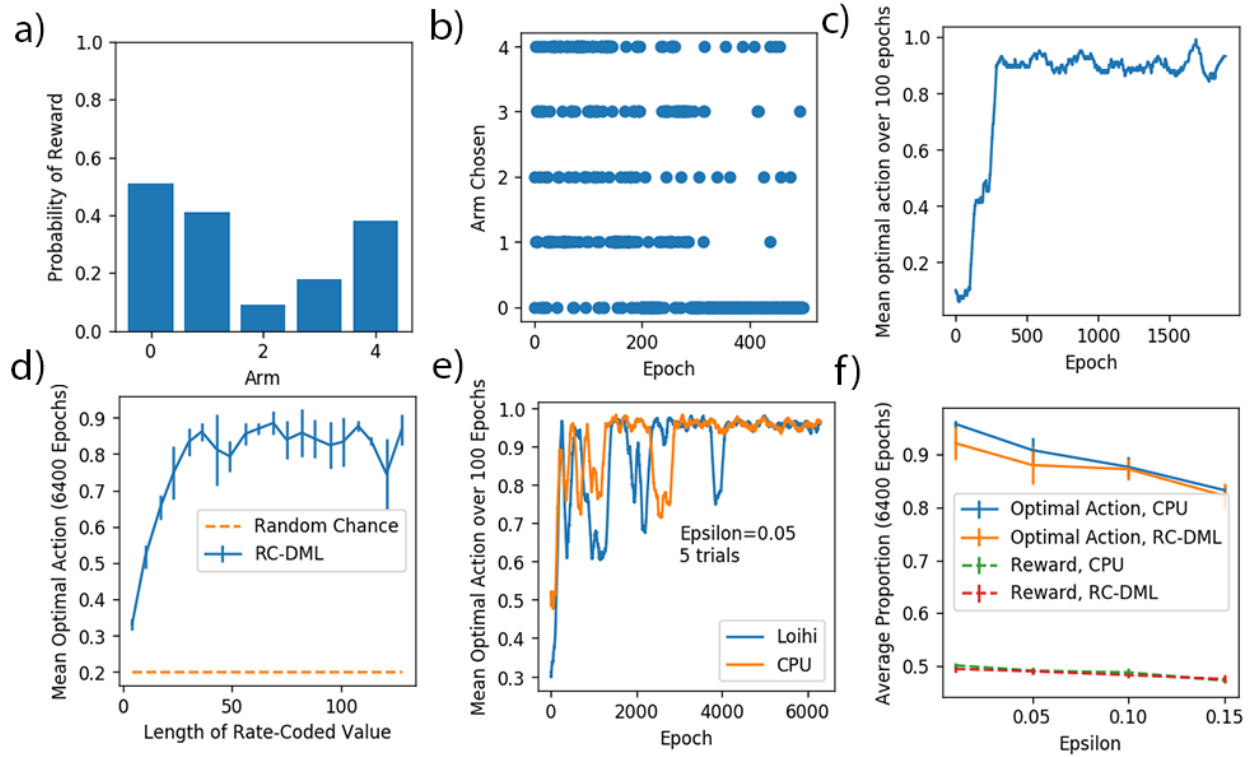

*Figure 4: Results from the RC-DML solution to the multi-arm bandit problem. A) The parameters of the Bernoulli distributions used to draw rewards after the pull of an arm. B) The choices of the RC-DML over 2000 epochs (arm pulls). C) The mean optimal action (MOA) of the RC-DML bandit, averaged over 100 epochs. D) The MOA of the bandit as the length of its rate coded representations is varied. Bars indicate standard deviation (n=5). E) Comparison of the MOA between the RC-DML executing on Loihi and a traditional CPU-based ε-greedy algorithm. F) Further comparisons between the RC-DML and CPU ε-greedy algorithms measuring MOA and cumulative reward as ε is varied. Bars indicate standard deviation (n=5).*

As long the length of the period used to rate-code values lies above approximately 40 steps, the neuromorphic RC-DML demonstrates learning performance for the MAB which is comparable to that of a traditional, non-spiking ε-greedy algorithm running on a CPU (Figure 4d, e). The CPU-based algorithm maintains a small performance lead in proportion of optimal actions and the mean average reward over the first 6400 epochs of learning over a variety of ε values (Figure 4f). This is due to the limited accuracy of value representation in the RC-DML due to its use of rate-coding, which later limits its performance at Blackjack. No such obstacles are present in the CPU-based algorithm which uses 64-bit floating point representations of value.

### Dynamic Maze

Navigation is a long-studied problem with many practical applications, such as robotic cleaners and self-driving vehicles. We focus on a dynamic maze task to evaluate the RC-DML's ability to learn navigation patterns. In this dynamic maze, a target location within a rectangular grid of points provides a reward. If the agent can reach this reward within a set number of moves, it is rewarded. During this time, the only information provided to the agent is its current location within the maze. It cannot 'see' walls, but infers their presence by detecting its location remains unchanged after attempting to move in a certain direction. If the agent fails to reach the goal within a set number of steps, it is punished and a new episode begins. The multi-state nature of navigation requires the RC-DML to demonstrate that it can accurately incorporate an episodic memory into its learning process. Additionally, the RC-DML here employs a greedy policy combined with a random starting location to force exploration.

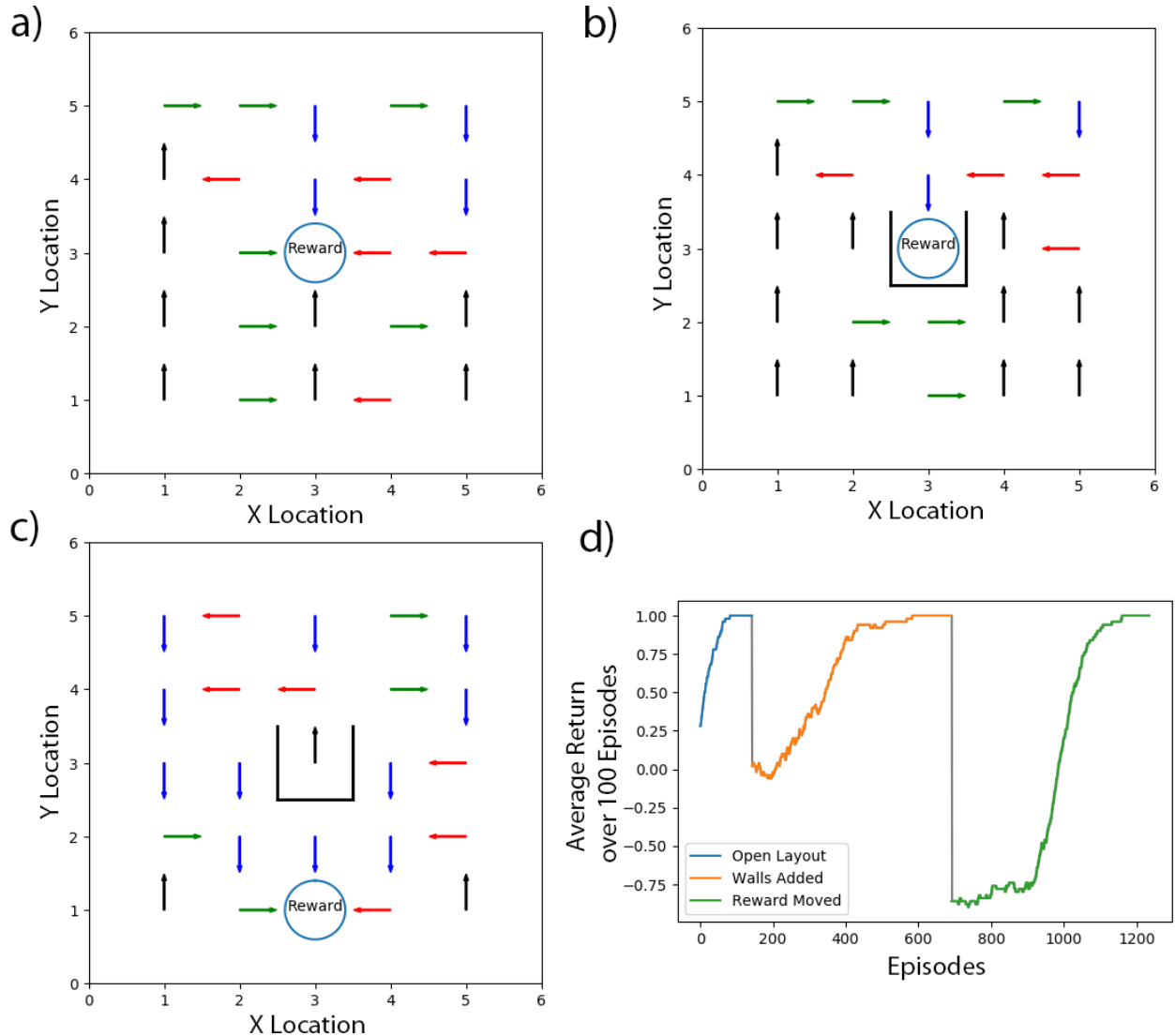

Figure 5: Results from the RC-DML applied to a navigation task. A) First, the agent must learn to navigate to a reward at the center of the rectangular arena. Arrows indicate the direction in which the policy will guide the agent. B) After it has learned this simple navigation task, walls are introduced which only allow the reward to be approached from the North; a new policy must be learned to overcome this change. C) Finally, the reward is moved from its location at the center to the bottom of the arena. Value estimates across the entire arena must be re-learned to form an optimal policy which overcomes this change. D) The average returns for the RC-DML agent over 100 epochs are plotted. This metric is calculated separately for each configuration of the arena.

Given the consistent nature of rewards, the RC-DML can quickly learn the location of the reward within a 5 by 5 grid and converge on a policy which navigates to it from all starting locations within the time limit of 8 steps (Figure 5a). When walls are introduced within the maze to block off the reward from 3 sides, the previous navigation policy conflicts with the new constraints imposed by the environment. But as the value estimates within the RC-DML are effectively created by a running average of previous returns, it gradually overcomes this faulty policy as the new experiences gained from the environment update value estimates and allow a better policy to emerge which correctly navigates to the reward (Figure 5b).

Finally, the reward is placed at a new location within the maze. This requires a greater shift in the navigational policy than when the previous reward location is blocked off, and the RC-DML takes a longer time to adjust to this new environment, but once again the optimal policy is reached with experience (Figure 5c,d).

### Blackjack

The card game Blackjack contains elements of random chance which makes it challenging to quickly converge to the optimal policy. In the variation of Blackjack we implement, the goal is to obtain a hand of cards with values summing as close as possible to 21 without going over. Each pip card (2-9) has a value equivalent to its number, face cards are worth 10 points, and the ace can count as either 1 or 11. To begin play, a card is dealt to both the dealer and the player. From there, the player chooses to either receive a new card from the dealer ('hit') or stay with their current sum ('stick'). The dealer then follows a fixed policy to draw cards until their sum is 17 or greater. If either the player is closer to 21 or the dealer goes over 21 during their turn, the player wins. If both player and dealer are equally close to 21, the game is a draw. Otherwise, the dealer wins and the player loses.

To solve this problem, the RC-DML again uses a greedy policy combined with exploring starts. The state of the player in blackjack is entirely determined by the current sum of the player's cards (player sum), whether this sum can be modified by counting an in-hand ace as either 1 or 11 without exceeding 21 (usable ace), and the value of the dealer's visible card (dealer showing). Cards are drawn from an infinite deck, therefore there is no hidden state within the dealing process to consider.

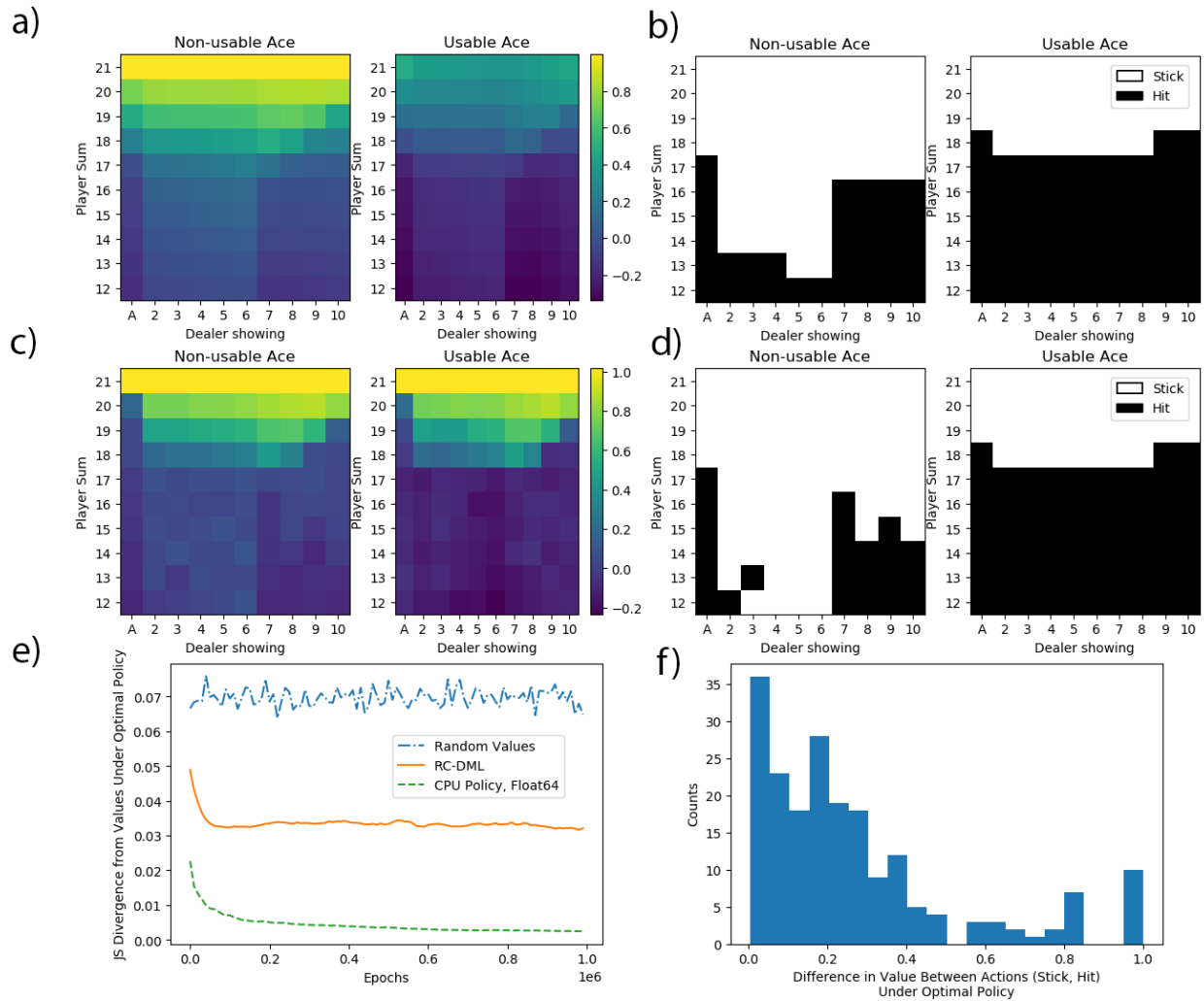

*Figure 6: Results from applying the RC-DML to playing Blackjack. A) Difference in the value between sticking & hitting under an optimal policy evaluated on CPU and (B) the optimal policy formed by applying a greedy choice to these values. C) Difference in the value between sticking & hitting after 1 million episodes of learning with the RC-DML on Loihi and (D) the policy formed from these values. E) Jensen-Shannon (JS) divergence between each agent's value representations and those under the optimal policy (taken after 2 billion episodes of learning on CPU). The coarse representation of values within the RC-DML limit its performance in this task and prevent it from reaching the optimal policy. However, its performance is still much better than random chance. F) The difference in policies between CPU and RC-DML corresponds to states where the difference in value between actions is small and requires fine representation.*

Although the RC-DML converges onto a policy which is quantifiably better than random chance and does not make obvious mistakes (such as drawing a new card when the player sum is already close to 20), a gap remains between its performance and that of an equivalent MC algorithm on CPU (Figure 6a-e). As previously mentioned, the limitations of rate-coding value estimates are the main source of this performance gap. Compared to the previous navigation problem, learning an optimal Blackjack policy requires much more fine-grained representations of value (Figure 6f).

# Discussion:

We have presented a novel framework for solving reinforcement learning (RL) problems using spiking neuromorphic hardware. This method implements critical elements and principles of the complimentary learning systems theory which was proposed to explain declarative memory learning in the biological brain. The method was successfully applied to three classic RL problems - Multi-Arm Bandit, Dynamic Maze, and Blackjack - using Intel's Loihi neuromorphic processor. We found that Loihi based implementation presents similar performance to the CPU based algorithms unless high precision was required to successfully learn a policy. While the active power consumption of a Loihi chip running the algorithm was much lower than that of a traditional CPU, for more complex problems, such as Blackjack, the faster execution rate of an equivalent Monte Carlo program on a CPU gave it an advantage in the overall energy consumption. The last limitation was found to be a result of rate-coding implementation of the information processing, that we used here due to its easy interpretability and functionality. Preliminary analysis revealed that using a different representation of information could succeed in making the proposed RL implementation competitive with traditional CPU solutions both in terms of learning capability and energy efficiency.

**Reinforcement learning in machine learning solutions**

Since its formal emergence in the mid-20th century, reinforcement learning (RL) has grown to provide a variety of techniques which can provably converge on optimal solutions for complex problems[6]. With the addition of modern deep-learning techniques, RL has surpassed human performance on a variety of problems, including the classic games Chess, Shogi, and Go[14]. Due to its ability to conquer these complex scenarios, RL has many possible applications in real-world problems including autonomous driving, process control, and interpreting biological data[15–17]. However, deploying RL in these and other edge-case situations requires that it be implemented in an efficient manner[1]. The desire for low-power implementations of RL, combined with the theory that neural circuits carry out some form of RL, has motivated several previous efforts towards implementing RL on neuromorphic platforms [6,18–20].

The fundamental problem which RL solves is providing an 'agent' in an environment a way of updating its internal models to choose an action at each state of the problem which leads to a maximized cumulative reward, or 'return.' Generally, this is done by iteratively improving the agent's value estimates, which are then applied to form a new policy to guide actions. Value estimates following the new policy are updated, and the process repeats. As long as this process satisfies the Bellman optimality principle, this process is guaranteed to converge to an optimal policy[6].

Implementing these mechanics on a neuromorphic platform gives rise to a number of challenges, such as how to represent and update values, how to update these values using local information, and how this program can be expressed in a manner which may be implemented on multiple architectures despite their underlying differences [2,3,10].

To address these challenges, we created a flexible program carrying out an RL strategy described by a high-level computational graph. In order to create this program, we focused on the requirements of an RL program and how they can be satisfied by a biologically-inspired structure. Once defined, we implemented this program on a neuromorphic platform in a modular manner, allowing it to address different problems with minimal changes from the user.

**Learning in biological brain and complimentary learning systems theory**

The mammalian brain contains analogs to the functions required for MC methods. Dedicated areas of the brain focus on maintaining robust internal representations of state, including physical location and proprioception[21,22]. It is believed that the hippocampus can fuse these complex representations of state over time, allowing animals to 'record' the paths they have taken in physical environment and play them back[23,24]. The short-term memories first formed in hippocampus can then be used to build more complex representations and strategies as they are integrated into the larger, long-term memory provided by the neocortex. The dual-memory theory[5] suggests that these two areas provide complementary functions to one another: the hippocampus provides a highly-plastic memory to learn from new experiences, and in order to prevent catastrophic forgetting the neocortex slowly integrates this new information into its already-existing representations [5].

In this work we implemented main elements of the complementary learning systems by utilizing separate short-term memory (STM) and long-term memory (LTM) modules complemented by decoder and encoder blocks.

The decoder's role is to interpret and preserve information from the environment. Viewing sensory information from the environment as a coded signal, the decoder's task is to extract everything from this signal relevant to the current state and store this internally for other modules to use (e.g. given an image of a chess board, evaluate the placement of the pieces and represent this state information in a manner meaningful to the downstream modules).

The STM is responsible for storing the agent's episodic memory. Given internal representations state, action, and reward the agent experiences during an episode, the STM must store this information and allow it to be 'replayed' when the episode has been completed and can be used to update value estimates. Consequently, the STM utilizes a highly plastic memory which updates with every action the agent takes.

One of the central pieces of any RL agent is its value estimator, which here must provide the return expected given the current state, next action, and policy. This information must be constructed via cumulative experiences and/or internal bootstrapping, with each additional experience building upon previous knowledge rather than overwriting it. This requires the value estimator to have a stable, long-term memory. Cumulative experiences are integrated into the LTM by replaying the STM (akin to sharp-wave ripples within the brain[24]), and selective replay within the LTM can potentially be used for bootstrapping. Ultimately, the value estimates stored in this module are responsible for constructing the agent's current policy, controlling its decisions.

The encoder carries out the final sub-step in an RL epoch, applying the current state and policy to choose an action. Additionally, this action may need to be transformed from an internal state into one which can act on the environment (e.g. the intent to move a piece in Chess must be translated into the physical act of picking up and moving an object). These tasks are carried out by the encoder, closing the loop of the RL process.

Having defined these modules and how they act in concert to perform RL, the challenge was to translate them from abstract concepts into realizable implementations which can be implemented on a real neuromorphic processor. In this work, we focused on demonstrating the architecture with a simple proof-of-concept implementation which can be easily interpreted. However, more advanced representations and other advances in neuromorphic computation can be integrated into the model in the future, and we posit this could lead to large gains in efficiency over the current implementation.

## Power Consumption

One of the key goals of neuromorphic hardware is to increase the efficiency of learning and inference in artificial systems[2]. However, the current implantation of the DML makes several trade-offs which reduce its energy efficiency. First, information is represented through rate-coding; this requires that each step of inference in a decision-making process be run over a several time steps in order to collect statistics on current spike rates[25]. Second, a dedicated spiking neural circuit exists in the LTM for every state-action value which must be estimated. These circuits run continually and in parallel, even when their information is not needed. In a problem such as Blackjack (with the largest state-action space tested of 400), this means much of the energy consumed to produce spikes is wasted, as only 0.5% of estimates being produced are relevant to the current state during each epoch. As a result, while the active power consumption of a Loihi chip running the RC-DML algorithm to solve Blackjack is much lower than that of a traditional CPU, the faster execution rate of an equivalent Monte Carlo program on CPU gives it an advantage in the amount of energy consumed per RL epoch (67 µJ/epoch on CPU vs. 1728 µJ/epoch on Loihi when using 64 rate-coding steps).

## Current Issues and Future Directions

Two key issues of the RC-DML (its uncompetitive power consumption and poor scalability) stem from the same underlying source: the rate-coded and tabular implementation of value estimates in the LTM. We argue that replacing this block with a different representation of information calculated through a function approximator which stores information passively in synapses would succeed in making the DML competitive with traditional CPU implementations, both in terms of learning capability and energy efficiency.

For instance, vector-symbolic architectures (VSAs) offer an alternative basis of representing information which could be leveraged within the DML[26]. Binary Spatter Codes (BSC) in particular provide a clear method towards replacing the rate-coding of information[27]. This would both increase the number of values a single population of neurons can represent and reduce the amount of time required to produce a representation to a single time-step. Potentially, this single change could allow the DML to outperform the CPU in per-epoch energy consumption, as running the current RC-DML architecture reduces its power consumption to 27 µJ/epoch when only a single computation step is executed per RL epoch.

A larger challenge facing neuromorphic implementations of the DML is an efficient and effective method for training function approximators such as deep neural networks. This advance was a key requirement for recent progress in RL[6], and equivalents within neuromorphic platforms must exist to enable state-of-the-art RL techniques on these platforms. However, advances have been made on this front and should be evaluated to be incorporated into future revisions of the DML on neuromorphic platforms which support them[10,11,28,29]. This would allow the architecture to scale to more complex problems with larger state-spaces.

## Conclusion

Reinforcement learning provides unique learning capabilities and its development has provided many landmark successes over the past decade. Therefore, it is crucial for neuromorphic systems to show that they are capable of RL techniques and can demonstrate advantages for these techniques over traditional hardware. In this work, we have demonstrated a flexible architecture for RL on neuromorphic hardware which was implemented and fully executed on the Intel Loihi platform. This rate-coded dual memory learner (RC-DML) was successfully able to learn policies to maximize reward received from a multi-arm bandit, navigate through a changing maze, and play the card game Blackjack. But while this shows that a neuromorphic architecture is currently capable of RL techniques, the current implementation's rate-coded and tabular approach to value representation makes it uncompetitive with traditional technologies. However, we believe that further advances from research into neuromorphic systems (such as value representation through vector-symbolic architectures and spike-based deep learning) can overcome this obstacle in future work to create a system which can match the performance of traditional approaches with greater energy efficiency.

# Materials & Methods

All spiking networks were developed using Python 3.5.2, the Intel NxSDK v0.9.5 - v0.9.9, and were executed on the Intel Loihi processor through the Intel Neuromorphic Research Community (INRC) cloud. Code for the RC-DML and each task is available online (https://github.com/wilkieolin/loihi_rl).

## Value Circuit

The Value Circuit (VC) is a small spiking circuit which provides the long-term memory and learning capability of the RC-DML. It accomplishes this task by taking an approximate moving average of sparse reward signals and representing this value with a continuous stream of rate-coded spikes.

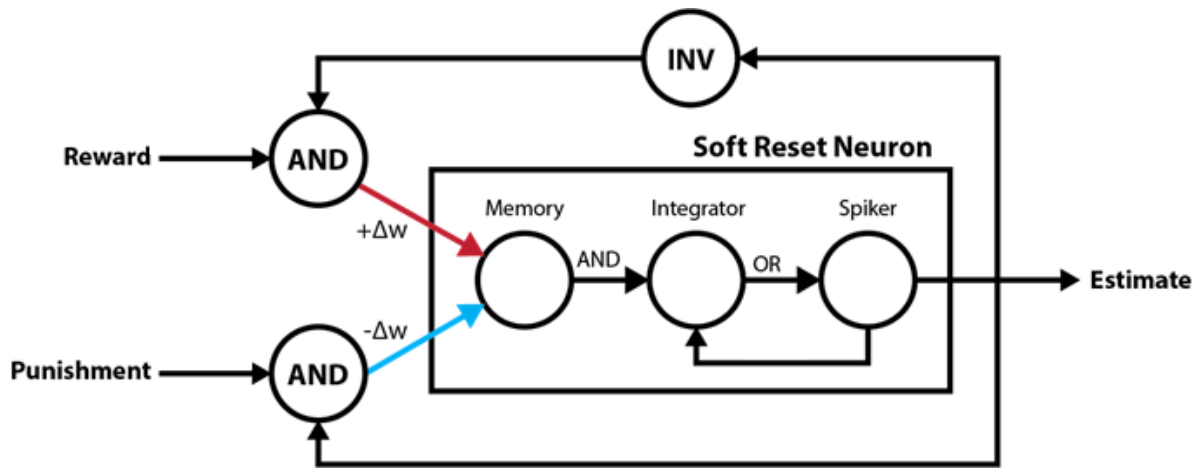

*Supplementary Figure 1: Illustration of the Value Circuit (VC). Reward signals have a chance to increase the firing rate of the soft-reset neuron, and punishment signals, a chance to decrease the firing rate. Through the use of feedback, these probabilities are manipulated to be equal to one another when the firing rate of the soft-reset neuron equals the proportion of reward spikes received over a memory period.*

The VC operates by using feedback to manipulate the probability its firing rate will change when a reward or punishment signal is received. The VC's output is given by the firing rate of a soft-reset integrate-and-fire (SRIF) neuron with no leakage; this 3-compartment neuron produces a regular firing rate when the charge in its memory compartment is held constant[25]. This firing rate is linearly proportional to the amount of charge in the memory compartment versus the integrator's firing threshold. In order to increase the firing rate, charge is added to the memory compartment through an excitatory synapse, and to decrease it, charge is removed through an inhibitory synapse.

However, if we wish to use this neuron to track the expectation of a reward, charge should not be modified with the arrival of every reward or punishment signal; this is clear if the neuron is already firing tonically or is quiescent. More generally, by using the spiking output of the SRIF in conjunction with an 'and' gate which only fires when spikes are present on all its inputs in a single time-step, punishment signals can be blocked or passed to make modifications to charge which are dependent on how close to the quiescent state the SRIF is. To create a similar condition for the reward signals, an inverter is added to the signal path which fires whenever an input is not present. This restricts reward signals, making them less likely to pass the closer the SRIF is to the tonic state. By combining these two boundaries, the probabilities that the SRIF's firing rate will increase or decrease become equal when the firing rate matches the proportion of reward signals out of reinforcement signals received over a finite period. This duration of this 'memory window' is affected by increasing or reducing the amount of charge injected into the SRIF's memory compartment via synapses or changing the SRIF firing threshold. The larger the ratio of injected charge to threshold, the shorter the effective memory window (Supplementary Figure 2).

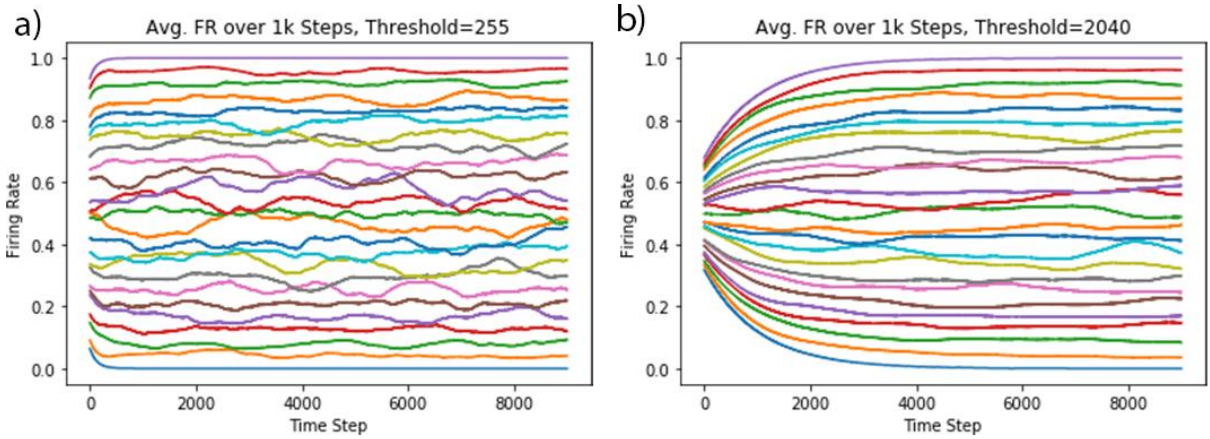

*Supplementary Figure 2: Firing Rates of value circuits (VCs) with different integrate-and-fire thresholds in the soft-reset integrate-and-fire (SRIF) neurons. A low threshold value leads to a shorter memory period (a) than a high threshold value (b), as more larger changes in charge are required to significantly alter the firing rate of the high-threshold VC.*

The convergence of the SRIF is shown formally by Eqns. 1-3 where $+\Delta q$ and $-\Delta q$ are increases or decreases (respectively) in SRIF charge, $s$ is the state of the SRIF's output, $\hat{r}$ is the SRIF's firing rate estimating the true parameter of Bernoulli-distributed reward $r$, and $reward/punishment$ indicates a sample of 1/0 from this distribution.

Eqn 1. $\quad p(+\Delta q) = p(s = 0|\hat{r}) \cdot p(reward = 1|r) = (1 - \hat{r})(r)$
Eqn 2. $\quad p(-\Delta q) = p(s = 1|\hat{r}) \cdot p(punishment = 1|r) = (\hat{r})(1 - r)$
Eqn 3. $\quad p(+\Delta q) = p(-\Delta q) \, iff. \, \hat{r} = r$

## Block Diagrams

Here we detail structures used to carry out the tasks required in each module of the RC-DML. These modules are implemented via spiking equivalents of common digital logic functions. Each arrow in the figure represents a stereotyped connectivity pattern between blocks which transfers information via spikes. Each block automatically scales to the dimensions required by the given problem, reducing programming burden which switching the architecture to new tasks. These figures follow closely to the actual code which implements them.

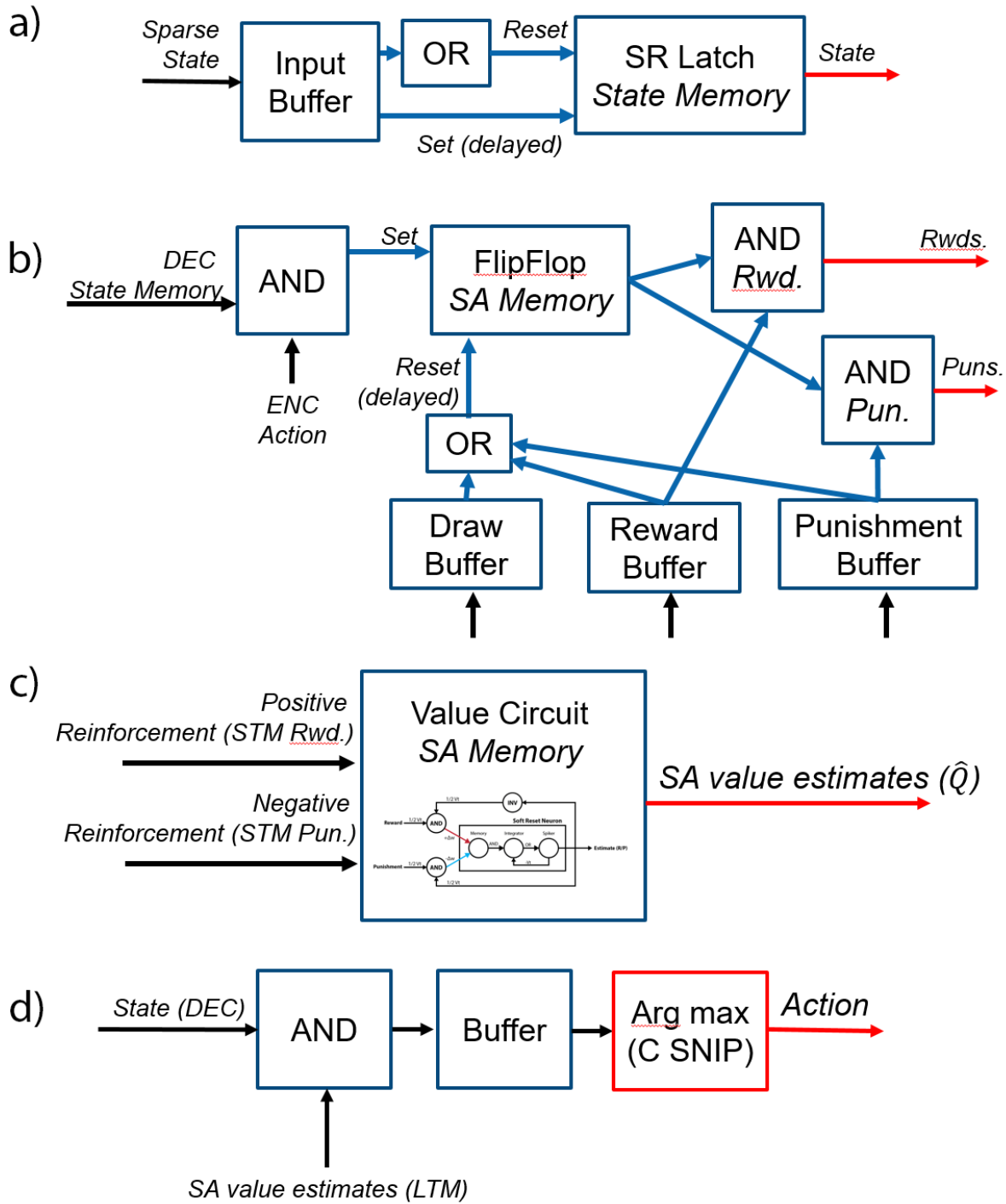

*Supplementary Figure 3: Block diagrams of the RC-DML modules. A) The decoder receives sparse, one-hot signals. Upon receiving a signal, it resets the memory bank and then sets the new current state, expanding it from a sparse signal in time to one which has a continuous availability to the rest of the system. B) The short-term memory stores the state-action pairs which have been encountered so far in an episode. This is done by forming an outer-product between current state and last action. When a new action arrives, its product is used to indicate this pair has been encountered. When a reinforcement signal arrives, state-action pairs which were encountered are used to signal the LTM to possibly change its values via rewards and punishments. The memory buffer is then reset. C) The LTM consists of a large bank of VCs which are detailed above. It receives reward/punishment signals which are distributed to appropriate state/action pairs from the STM. D) The encoder reads from the*

*LTM reward estimates corresponding to the current state. It accumulates these spikes in a buffer which is then read out by a C subroutine on the onboard microprocessor to select the highest value.*

## Power Estimates

Power estimates for CPU tasks were estimated by utilizing the Intel SoC Watch Energy Analysis profiler on a Windows 10-based system with an Intel i7-4710HQ processor (22 nm node). Active power used to calculate the optimal Blackjack policy was measured by running the included RL program 5 times and subtracting the system's baseline power usage (also measured over 5 independent trials).

Power estimates on the Loihi system was measured by using energy probe tools included in the NxSDK 0.9.9 toolkit. Programs were run on the ncl-ext-ghrd-01 system available on the Intel Research cloud platform. Energy consumption for the RC-DML learning Blackjack was monitored over 128,000 computation time steps and averaged.

## Funding

This work was supported by the Intel (00018020-001) and Lifelong Learning Machines program from DARPA/MTO (HR0011-18-2-0021).

## Acknowledgments

We would like to thank the members of the Intel Neuromorphic Research Lab for their support during our software development process.